\def\year{2021}\relax
\documentclass[letterpaper]{article} 
\usepackage{aaai21}  
\usepackage{times}  
\usepackage{helvet} 
\usepackage{courier}  
\usepackage[hyphens]{url}  
\usepackage{graphicx} 
\usepackage{natbib}  
\usepackage{caption} 
\usepackage{multirow,tabularx}
\usepackage{booktabs} 
\usepackage{arydshln}
\usepackage{textcomp}
\usepackage{amsmath}
\usepackage{soul}
\usepackage{pifont}
\usepackage[table]{xcolor}

\makeatletter
\def\adl@drawiv#1#2#3{%
        \hskip.5\tabcolsep
        \xleaders#3{#2.5\@tempdimb #1{1}#2.5\@tempdimb}%
                #2\z@ plus1fil minus1fil\relax
        \hskip.5\tabcolsep}
\newcommand{\cdashlinelr}[1]{%
  \noalign{\vskip\aboverulesep
           \global\let\@dashdrawstore\adl@draw
           \global\let\adl@draw\adl@drawiv}
  \cdashline{#1}
  \noalign{\global\let\adl@draw\@dashdrawstore
           \vskip\belowrulesep}}
\makeatother

\title{Multi-Task Learning for Situated Multi-Domain End-to-End Dialogue Systems}
\author {
     
Po-Nien Kung, 
     Chung-Cheng Chang, 
     Tse-Hsuan Yang, 
     Hsin-Kai Hsu, 
     Yu-Jia Liou, \\
     Yun-Nung Chen \\
}
\affiliations {
Department of Computer Science and Information Engineering \\
National Taiwan University, Taipei, Taiwan \\
\{b06902012, b06902118, b06902032, r07922079, b06902008\}@csie.ntu.edu.tw, \\
 y.v.chen@ieee.org\\
}

\begin{document}

\maketitle  

\begin{abstract}
Task-oriented dialogue systems have been a promising area in the NLP field. Previous work \cite{hosseini2020simple, Wolf2019HuggingFacesTS, zhao2016endtoend, budzianowski2019hello} showed the effectiveness of using a single GPT-2 based model to predict belief states and responses via causal language modeling. In this paper, we leverage multi-task learning techniques to train a GPT-2 based model on a more challenging dataset~\cite{moon2020situated, crook2019simmc} with multiple domains, multiple modalities, and more diversity in output formats. 
 Using only a single model, our method achieves better performance on all sub-tasks, across domains, compared to task and domain-specific models. Furthermore, we evaluated several proposed strategies for GPT-2 based dialogue systems with comprehensive ablation studies, showing that all techniques can further improve the performance.
\end{abstract}

\begin{table*}[t]
\centering
\begin{tabular}{lp{14cm}}
\toprule
\textbf{User utterance} & \it Yes I think that will work. What is it's size?\\
\bf Visual objects& OBJECT\_0 : pos focus color [`Beige', `Brown'] class\_name Area Rugs decor\_style [`Modern']] \\
\bf Belief state & \textbf{DA:INFORM:PREFER:FURNITURE} [ furniture-O = OBJECT\_0, furniture-attentionOn = that ] \textbf{DA:ASK:GET:FURNITURE.dimensions} [ furniture-O = OBJECT\_0 ] \\
\bf Action & specify info [ matches = dimensions ] \\
\bf System response & \it This item is 91.77 inches wide and 91.63 inches deep \\
\bottomrule
\end{tabular}
\caption{An example of the different modalities from one turn of a dialog in the SIMMC-furniture dataset. The \textbf{Bold} part of the state is the dialog act, and the rest is slots.
}
\label{tab:example}
\end{table*}

\begin{table*}
    \centering
\begin{tabular}{|l|cc|ccc|ccc|ccc|}
        \hline
        \multirow{2}{*}{\bf Dataset} & \multirow{2}{*}{\bf \#Actions } & \multirow{2}{*}{\bf \#Attributes} & \multicolumn{3}{c|}{\textbf{Train set}} & \multicolumn{3}{c|}{\textbf{Dev set}} & \multicolumn{3}{c|}{\textbf{Devtest set}} \\
        & & & 
        \textbf{nums} & $\overline{\textbf{T}}$ &  $\overline{\textbf{L}}$ & 
        \textbf{nums} &  $\overline{\textbf{T}}$ &  $\overline{\textbf{L}}$ & 
        \textbf{nums} &  $\overline{\textbf{T}}$ &  $\overline{\textbf{L}}$ \\
        \hline
        Fashion & 5 & 7 & 21196 & 5.4 & 140.5 & 3513 & 5.4 & 139.8 & 5397 & 5.5 & 141.7 \\
        Furniture & 7 & 60 & 29213 & 7.6 & 115.8 & 4827& 7.5 & 112.7 & 7355& 7.7 & 115.5\\
        \hline
    \end{tabular}
    \caption{The statistics of the 2 datasets we used.
    $\overline{T}$ stand for the average turns in that set. 
    $\overline{L}$ is the average number of tokens per example fed into the model, where we only use the past 2 turns as context.
    }
    \label{tab:dataset}
\end{table*}

\begin{figure*}[t]
  \includegraphics[width=\linewidth]{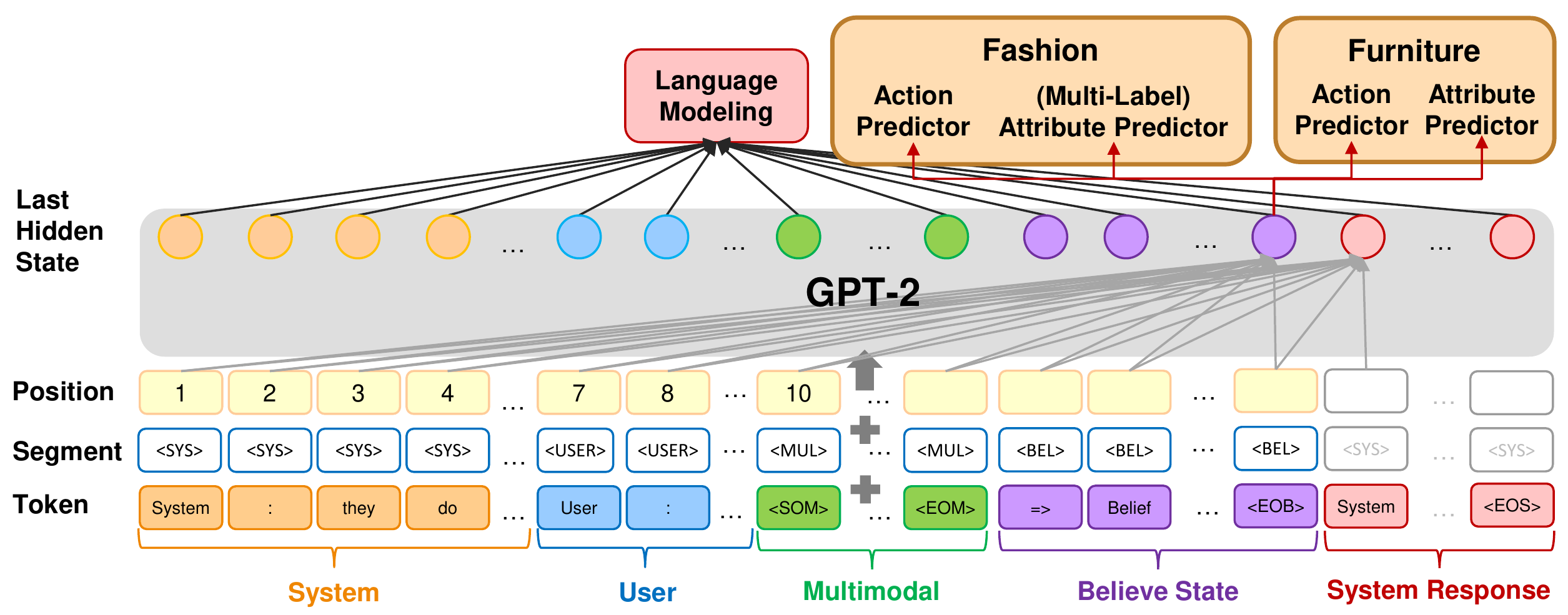}
  \caption{The proposed model architecture and multi-task training objectives.}
  \label{fig:model}
\end{figure*}

\section{Introduction}
A conversational agent, or a dialogue system is a program which interprets and responds to statements made by users in natural language. Most dialogue systems contain the capabilities of natural language understanding (NLU) and natural language generation (NLG). Dialogue systems can usually be clustered into two categories, Chit-Chat-Oriented dialogue system and Task-Oriented dialogue (TOD) system.  The former aims to generate meaningful and diverse response for open-domain conversations, and focus on language understanding and generation. This system is usually trained on large-scale data collected on the internet or social media. For the Task-Oriented dialogue system, its main objective is to understand the conversation and solve specific tasks. This system is usually decomposed into three tasks, NLU, NLG, and Action Prediction. Using Dialogue State Tracking system, Language Generator and Action predictor to track belief-states, generate responses and predict actions, respectively. 
Traditional TOD systems are constructed with several modules, each independently trained to tackle a different sub-task. Though this concept of the traditional TOD system seems to be intuitive, there still exist some limitations: 1) All the separate components do their job independently, while sometimes the tasks knowledge are shareable. 2) Due to the growing complexity of the TOD tasks, which contains more sub-tasks and complicated conversations, the standalone structure for each sub-tasks requires considerable human effort on structure design, and the model complexity grows. Based on the above concerns, End-to-End Task-Oriented DST has been proposed \cite{chao2019bertdst, peng2020soloist, ham-etal-2020-end, budzianowski2019hello}. They deal with the TOD problems using simplified structures, which usually consist of a transformer-based pre-trained language model for the NLU and NLG tasks, regarding both the belief state tracking and response generation as generation tasks, and solve these problems in a seq-to-seq fashion. By applying this structure, most tasks can be encoded as text generation task, which effectively simplifies the model structure, and by using a single model for all tasks, the learnt knowledge is shared among all sub-tasks. 

Recently, multi-task learning has been a popular approach \cite{liu2019multitask, raffel2020exploring}. The core idea of multi-task learning is knowledge transfer, which implies a shared representation \cite{rastogietal2018multi}. Usually, multi-task learning consists of two main components. One is a shared network for the shared representation, the other one is task-specific sub-networks for individual tasks. In End-to-End Task-Oriented DST, besides utilizing language modeling tasks for belief state tracking and response generation, it sometimes requires separate network structure for API prediction.
By applying the technique of iterative training in multi-task learning \cite{gulyaev2020goal}, we are able to use a shared network to encode the states information and dialogue history and use separate task-oriented predictor to do the API predictions. This way, we are able to reduce the number of network parameters by sharing encoder and improve the performances by sharing sub-task knowledge, when applied to different down-stream tasks simultaneously.

Multi-Domain DST has been a promising research area in the near ten years. Multiple multi-domain DST datasets have been released\cite{budzianowski2018multiwoz,rastogi2020towards}. In previous researches, \citet{zeng2020multi} and \citet{9053248} investigate the strength using state graphs for multi-domain training, which effectively connect the information between different domains. Later \citet{zeng2020multi, hosseini2020simple, budzianowski2019hello} proposed a simplified method by using a pre-trained language model as a shared encoder to jointly learn multiple domains data, which can predict multi-domains data more precisely using a much simpler model structure.

Nowadays, more and more virtual digital assistants have been used and those assistants are expected to process not only the text but multi-modal information.
Given multi-modal information, the assistants should give appropriate response, or even execute relevant actions.
To this end, \citet{moon2020situated} proposed the Situated and Interactive Multimodal Conversations (SIMMC) datasets as a starting point in this new research direction, where a dialog system needs to assist humans online shopping. This dataset contains visual objects, user and system utterance and API information from two different but correlated domains. To deal with this challenge, we apply Multi-Task Learning to learn an End-to-End Multi-Domain DST system. By iterative training multiple domains and multiple tasks data, we are able to train a share encoder, with corresponding Task-Oriented Predictors (for API) and LM Head (for belief and response), to predict in both domain with performance improvement.

The contributions of this work are summarized as follows:
\begin{itemize}
  \item We proposed a simple model structure to predict all sub-tasks and both domains using one model. The proposed model outperforms the baseline models with a great margin in all sub-tasks.
  \item We experiment on various forms of input representation and training strategies in the generation-based End-to-End DST system, and most of them shows improvement on the performance.
\end{itemize}

\begin{table*}
\centering
\begin{tabular}{ |l|p{1.4cm}|p{11cm}|  }
\hline
\multicolumn{2}{|l|}{\bf Strategy} & \bf Input\\
\hline\hline
\multirow{2}{*}{Split Belief Intent} & Origin &=\textgreater Belief State : DA:ASK:GET:FURNITURE.dimensions  \\ 
\cline{2-3}
& Modified & =\textgreater Belief State : intent ask get furniture dimensions
\\
\hline
\multirow{4}{*}{Add API Actions} & \multirow{2}{*}{Origin} & System : This is our Hedon Kitchen Island with Stainless Steel Top. It features a natural wood countertop.  \\ 
\cline{2-3}
& \multirow{2}{*}{Modified} & System :  $<$searchfurniture$>$ This is our Hedon Kitchen Island with Stainless Steel Top. It features a natural wood countertop.
\\
\hline
\multirow{11}{*}{Mask History Loss} & \multirow{11}{*}{Modified} & \textit{System : Can you see now? User : no I cannot.  can you tell me about them?  $<$SOM$>$ OBJECT\_0 : pos left color [ White ] class\_name Kitchen Islands decor\_style [ Rustic Sophisticated ] OBJECT\_1 : pos center color [ White ] class\_name Kitchen Islands decor\_style [ Traditional Modern ]  $<$EOM$>$  System : This is our Hedon Kitchen Island with Stainless Steel Top. It featuresa natural wood countertop.} User : and what are the dimensions?  $<$SOM$>$ OBJECT\_0 : pos left color [ White ] class\_name Kitchen Islands decor\_style [ Rustic Sophisticated ] OBJECT\_1 : pos center color [ White ] class\_name Kitchen Islands decor\_style [ Traditional Modern ]  $<$EOM$>$ \hfill $=>$  Belief State : DA:ASK:GET:FURNITURE.dimensions [  ]  $<$EOB$>$  The width is 52 inches, depth 18 inches, and height is 36 inches.  $<$EOS$>$
\\
\hline
\end{tabular}
\caption{The examples of the modified inputs and masking history loss. For the \textbf{Split Belief Intent}, we show the original and modified belief states. For \textbf{Add API actions to Inputs}, one of the system response is shown. Noted that the actions are also added to the current turn system response (to be predicted). In the \textbf{Mask History Loss}  row, we represent the masked part(ignored by the loss function in the training phase) in the input with \textit{italic text}.}
    \label{tab:prop_features}
\end{table*}

\section{Dataset}\label{sec:Dataset}

We use the SIMMC (Situated Interactive Multi-Modal Conversations) described in \citet{moon2020situated}, namely SIMMC-Furniture (VR) and SIMMC-Fashion (Image) datasets. Both datasets consist of multi-turn multi-modal dialogues. Unlike most multi-modal datasets, all multi-modal information are provided in the text format.

Each turn contains a User utterance (U), a System/assistance utterance (S), a system Action (A), Visual objects (the Visual modality) (V), and a Belief State (B). Note that in this dataset, the Belief State is not accumulated across turns, i.e. the Belief State of one turn is the annotated Intent (and slots) of User utterance of that turn. An example is given in Table 1. More details can be found in \citet{moon2020situated}.

In the SIMMC dataset, both Furniture and Fashion domains include three sub-tasks: \textbf{Multimodal Action Prediction}, \textbf{Multimodal Dialog Response Generation \& Retrieval} and \textbf{Multimodal Dialog State Tracking (MM-DST)}. The detail of these three sub-tasks is described in subsection \ref{ssec:evaluation} and \citet{moon2020situated}.

\section{Proposed Approach}\label{sec:Model-Structure}
In this section, we discuss some proposed approaches that we've tried to improve the model performance. This includes changing the model structure and modify the input representation. First, we will go through the two baseline models proposed by Facebook AI Research team \cite{moon2020situated}. One is designed to predict both API and system responses. Another is built to predict belief states and system responses using a GPT-2 \cite{radford2019language} model with a linear language-modeling head.

\subsection{Baseline models}
In DSTC9~\cite{gunasekara2020overview} SIMMC track4\footnote{The DSTC9 SIMMC challenge page link is here. \url{https://github.com/facebookresearch/simmc}}, \citet{moon2020situated} released a baseline model with a complex architecture, containing multiple components including \textbf{Utterance \& History Encoder}, \textbf{Multi-modal Fusion} , \textbf{Action Predictor} and \textbf{Response Generator}. This model is able to predict API action, API attributes and system response.
Another model they proposed is a GPT-2 based End-to-End DST model, designed to predict the belief states and response.
This model uses a GPT-2 model with a language modeling (LM) head to auto-regressively generate belief states first and then generate system response, using the history/context as prompt.
Note that both baseline models can't handle all sub-tasks (API, Response, Belief States) and need to be trained on the Furniture and Fashion datasets separately.
In this paper, we propose a multi-domain multi-task end-to-end training strategy. Using only one model, we are able to achieve higher performance on all sub-tasks in both domains.

\subsection{Proposed Model}
Following the GPT-2 based baseline model, we extend it to predict all sub-tasks in both Furniture and Fashion domains using only one model. We use a pretrained GPT-2 \cite{radford2019language} model with a LM head as the main structure to predict \textbf{Multimodal Dialog Response Generation \& Retrieval} and \textbf{Multimodal Dialog State Tracking (MM-DST)}. For the \textbf{Multimodal Action Prediction} subtask, it requires API actions and API attributes prediction. Thus we added four classifiers after the GPT-2 Model to predict \textbf{Furniture API action}, \textbf{Furniture API attribute}, \textbf{Fashion API action} and \textbf{Fashion API attribute}, respectively.
\paragraph{Task-Oriented Predictors}
Since API action and API attributes are classification tasks\footnote{For Fashion API attributes prediction, it is a multi-label classification task. Others are multi-class classification tasks}, we use a 3 layer multi-layer perceptron (MLP) classifier to predict each classification task. The first 2 layers each reduces the dimension by 2, and is followed by ReLU activation. Noted that all of these Task-Oriented Predictors use a single round hidden states output by the main GPT-2 model as input. In our proposed model, we use the hidden state of the token "\textless EOB\textgreater" as the input. In practical, after the main LM layer output token "\textless EOB\textgreater", we use the current round hidden state output from GPT-2 model as the input of all the Task-Oriented Predictors. We would like to note that although it is possible to predict API auto-regressively in the sequence-to-sequence fashion, this approach is more flexible in that we can use arbitrary task-specific layers for different kinds of tasks \cite{liu2019multitask}. This is further discussed in subsection \ref{ssec:text_gen}.

\subsection{Input Representation}\label{ssec:input}
We further discuss the input representation in our proposed model. The input structure is based on the GPT-2 Baseline Model. We also tried some methods to modify inputs intended to improve the performance.

\paragraph{Representing Visual Information}
For both Furniture and Fashion domain, the visual information (multi-modal data) are the descriptions of items currently visible to the user and assistant. The descriptions may include the relative position, style, color, and class of the items\cite{crook2019simmc}\cite{moon2020situated}. Since the visual information has been converted to text information, we follow the same procedure as baseline to convert this information stored in JSON format into shorter text.

\paragraph{Representing History}
Following the baseline, we construct the input sequence using \textbf{History} and \textbf{Belief + Response}. The \textbf{History} contains $T$ turns of \underline{System Response + User utterance + Multi-Modal Info}, each of them has respective delimiters to separate different segments. The turns number $T$ here is set to 2 for all of our experiments. The effects of using different $T$ number is further discussed in the section \ref{sec:discuss}. As the \textbf{Belief + Response}, it uses "=\textgreater Belief States : " as prompt and Belief State ends with "\textless EOB\textgreater". The response starts after the "\textless EOB\textgreater" and ends with a "\textless EOS\textgreater". For an example input sequence, please refer to \textbf{Mask History Loss} parts in the Table \ref{tab:prop_features}.

\paragraph{Representing Belief State}
Belief states include intents and corresponding slot key-value pairs. In our proposed model, it is transformed into text. The model generates them auto-regressively. Different from \cite{moon2020situated} which adds all intents from training set  into vocabulary as special tokens, we lowercase and split the intent tokens by ":". This enables our model to predict each intent as a combination of existing tokens with pre-trained embeddings and learn the correlation information between each intent. We refer to this method as Split Intent (SI). The example of split belief states can be found in Table \ref{tab:prop_features}.

\paragraph{Segment Embedding}
In the original input representation, we use delimiters for the model to better differentiate between user utterance, system response, belief state, and multi-modal information.
To further help model to distinguish different segments, we follow previous works \cite{budzianowski2019hello, ham-etal-2020-end} and add segment embedding as extra input. We create special tokens such as "\textless SEG\_SYS\textgreater","\textless SEG\_USER\textgreater", "\textless SEG\_BEL\textgreater", "\textless SEG\_MUL \textgreater" for different parts in the input. The final input embedding become \textbf{Token Embedding + Segment Embedding + Positional Embedding}.

\paragraph{Add API actions to Inputs}
In the SIMMC dataset, it is easy to observe that the system responses are highly related to the current turn API action. The SIMMC challenge also allows us to use the ground truth API for response generation during inference\footnote{Challenge description \url{https://github.com/facebookresearch/simmc/blob/master/TASK_INPUTS.md}}. Therefore, to generate better system responses, we additionally inject the API action information after each \textbf{System :} delimiters in the inputs, including the current turn response. 
By giving the model information about current turn API action, the model is expected to generate a more precise response. The example of adding actions to context can be found in Table \ref{tab:prop_features}.

\section{Training Strategy}\label{sec:training_strategy}
We further discuss the training strategy and their implementation details in this section. To train our proposed model on both Furniture and Fashion domains using one model, we add a domain-specific token in each input sequence and merge the 2 datasets together. For the API prediction, since it is different from the original LM task, we apply the iterative training technique, where we only calculate the loss for one task for each iteration. For the cross entropy loss calculation in LM task, we also tried to train our model with history masked, to avoid negative transfer from learning to predict the history. We now explain each of our modifications in detail.

\subsection{Merge Domain Data}
SIMMC datasets include two different domains -- Furniture and Fashion. 
These two datasets are both designed for online shopping scenario, with similar but different API actions and belief states formats. We adapt two domain information together by applying the multi-task learning technique along with the domain indicator, which is shown to be important for Multi-Domain Training (MDT) \cite{joshi2012multi}. To merge two datasets, we first add domain indicators "\textless FURN\textgreater" and  "\textless FASH\textgreater" at the start of the input context, for furniture and fashion data, respectively. After the domain indicators are added, we simply merge the training data of these two datasets and iteratively train our model.

\subsection{Multi-Task Learning} \label{ssec:Multi-Task-Learning}
In our proposed model, we extend the original GPT-2 LM head model with task-specific predictors, to predict the API action and attributes. We follow \citet{liu2019multitask} and iteratively sample one task and use the corresponding loss to optimize the whole model.


\subsection{Mask History Loss}
In our proposed model, we use causal language modeling to train the model to predict belief states and system response. In the original settings proposed by \cite{moon2020situated}, they calculate the LM loss over the entire input sequence, which means the model will learn to predict all the input sequences, including user utterance and visual context.
We observed that in this loss formulation, the model has more chance to predict the tokens in earlier turns, compared to the later turns, if we use a larger turn number $T$, as we wanted to explore the effect of turn number $T$ on performance. As dicsussed in \ref{sec:discuss}.
To avoid over-weighting earlier turns when the turn number $T$ is large, we mask out the loss for the \textbf{History} part except the current turn user utterance. And observed an improvement in performance even when $T = 2$. For an example of masked inputs, see Table \ref{tab:prop_features}.

\begin{table*}
\rowcolors{3}{Wheat1}{}
\begin{tabularx}{\textwidth}{|p{2.5cm}||X|X|X|X|X|X|X|X|X|X|X|X|}
\hline
&\multicolumn{3}{|c|}{Sub-task 1 API} &  \multicolumn{6}{|c|}{Sub-task 2 Response Generation \& Retrieval} & \multicolumn{3}{|c|}{Sub-task 3 Belief State} \\
\hline
Model/Metrics& Act. Acc. & Attr. F1. & Act Per.$\downarrow$ & BLEU & r1 & r5 & r10 & Mean$\downarrow$ & MRR & Intent F1& Slot F1& joint \\
\hline
\hline
Baseline (HAE) & 79.7& 53.6& 1.70& 0.075 & 12.9& 28.9 & 38.4 &31.0 &0.218 & X & X & X \\
Baseline (HRE) & 80.0& 54.7& 1.66& 0.075&13.8 &30.5 &40.2 &30.0 &0.229 & X & X & X \\
Baseline (MN) &79.2 &53.3 &1.71 &0.084 &15.3 &31.8 &42.2  &29.1 &0.244 & X & X & X \\
Baseline (T-HAE) &78.4 &53.6 &1.83 &0.044 &8.5 &20.3 &28.9 &37.9 &0.516 & X & X & X \\
Baseline (GPT2)\textsuperscript{\ding{104}} & X & X & X & 0.095 & 17.6 & 39.4 & 51.9 & 21.6 & 0.286 & 80.9 & 77.2 & 62.4 \\

\hline
\multicolumn{13}{|l|}{\textbf{\textit{\large Modifying Inputs}}}\\
Prop. w/o SI\textsuperscript{\ding{61}} & 80.2 & \colorbox{red!25}{67.3} & \colorbox{green!25}{1.80} & 0.114 & 21.1 & \colorbox{green!25}{44.5} & 56.9 & \colorbox{green!25}{19.2} & 0.325 & \colorbox{red!25}{82.8} & 79.3 & \colorbox{red!25}{65.2} \\
Prop. w/o SE\textsuperscript{\ding{61}} &80.1 & 68.7 & \colorbox{red!25}{1.84} & \colorbox{green!25}{0.118} & \colorbox{green!25}{21.8} & \colorbox{green!25}{44.7} & \colorbox{green!25}{57.3} & \colorbox{green!25}{19.2} & 0.331 & 83.0 & 79.5 & 65.5 \\
Prop. w/o AA\textsuperscript{\ding{61}} & \colorbox{red!25}{79.8} & 68.3 & \colorbox{red!25}{2.23} & \colorbox{red!25}{0.107} & \colorbox{red!25}{20.2} & \colorbox{red!25}{42.1} & \colorbox{red!25}{53.8} & \colorbox{red!25}{21.0} & \colorbox{red!25}{0.312} & \colorbox{green!25}{83.3} &\colorbox{green!25}{79.9} & \colorbox{green!25}{65.9} \\
\hline
\multicolumn{13}{|l|}{\textbf{\textit{\large Training Strategy}}}\\
Prop. w/o MD\textsuperscript{\ding{61}} & 79.92 & \colorbox{red!25}{66.13} & \colorbox{red!25}{1.86} & \colorbox{green!25}{0.118} & 20.8 & \colorbox{green!25}{44.6} & 56.5 & \colorbox{green!25}{19.5} & 0.324 & \colorbox{red!25}{82.4} & 79.4 & \colorbox{red!25}{64.8} \\
Prop. w/o MT\textsuperscript{\ding{104}} &X&X&X& 0.114 & 20.6 & \colorbox{red!25}{43.7} & 56.4 & 19.7 & \colorbox{red!25}{0.320} & 83.2 & 79.7 & 65.8  \\
Prop. w/o MHL\textsuperscript{\ding{61}} & 80.1 & \colorbox{red!25}{66.9} & 1.81 & 0.113 & 20.7 & 43.9 & 56.3 & 19.6 & 0.321 & 82.9 & 79.4 & 65.5 \\
\hline
Prop.\textsuperscript{\ding{61}} & 80.21 & 69.18 & 1.81 & 0.115 & 20.9 & 43.9 & 56.5 & 19.8 & 0.323 & 83.05 & 79.43 & 65.5 \\
SD(Prop.) & 0.3932 & 0.9829 & 0.0096 & 0.0017 & 0.33 & 0.18 & 0.44 & 0.2828 & 0.0029 & 0.2082 & 0.2986 & 0.2986 \\
\hline
\end{tabularx}

\caption{Results of the SIMMC-Furniture dataset. For the evaluation scores, only the \textbf{Act Per.} and \textbf{Mean} is the lower the better. We train our proposed full model with four different random seeds, the mean scores and standard deviations of these four models are shown in \textbf{Prop.} and \textbf{SD(Prop.)}, respectively. For the ablation models, each label after "w\textbackslash o" stands for a proposed feature or a training strategy. (SI \textrightarrow Split Intent, SE \textrightarrow Segment Embedding, AA \textrightarrow Add Action to Inputs, MD \textrightarrow Multi-Domain, MT \textrightarrow Multi-Task, MHL \textrightarrow Mask History Loss). The highlighted scores of the ablation models represent the performance drop (red)/improvement (green) compare to \textbf{Prop.}. If the scores of an ablation model is one standard deviation (\textbf{SD(Prop.)}) worse than the \textbf{Prop.}, the score is highlighted in red. For the performance improvement, we highlight the scores in green. All model names with \ding{61} are trained without multi-task for 6 epochs and further multi-task training 20 epochs, the best dev-test performance happened at 18th\texttildelow22th epochs. The model names with \ding{104} are trained without multi-task for 10 epochs, the best epochs happened at the 8th epoch.}
\label{tab:result_furniture}
\end{table*}

\begin{table*}
\begin{tabularx}{\textwidth}{|p{2.5cm}||X|X|X|X|X|X|X|X|X|X|X|X|}
\hline
&\multicolumn{3}{|c|}{Sub-task 1 API} &  \multicolumn{6}{|c|}{Sub-task 2 Response Generation \& Retrieval} & \multicolumn{3}{|c|}{Sub-task 3 Belief State} \\
\hline
Model/Metrics& Act. Acc. & Attr. F1. & Act Per.$\downarrow$ & BLEU & r1 & r5 & r10 & Mean$\downarrow$ & MRR & Intent F1& Slot F1& joint \\
\hline
\hline
Baseline (HAE) &81.0 &60.2 & 1.75 & 0.059&10.5 &25.3 &34.1 &33.5 &0.190 & -- & X & X \\
Baseline (HRE) & 81.9 & 62.1 & 1.76 & 0.079&16.3 &33.1 &41.7  &27.4 &0.253 & X & X & X \\
Baseline (MN) & 81.6& 61.6& 1.74& 0.065&16.1 &31.0 &39.4 &29.3 &0.245 & X & X & X \\
Baseline (T-HAE) & 81.4 & 62.1& 1.78 &0.051 &10.3 &23.2 &31.1&37.1 &0.178 & X & X & X \\
Baseline (GPT2)\textsuperscript{\ding{104}} & X & X & X &0.108 & 16.7 & 40.0 & 54.2 & 20.4 & 0.283 & 69.2& 68.9 &46.3  \\
\hline
\multicolumn{13}{|l|}{\textbf{\textit{\large Modifying Inputs}}}\\
Prop. w/o SI\textsuperscript{\ding{61}} & 85.38 & 69.71 & 1.95 & 0.135 & 22.4 & \colorbox{red!25}{47.3} & 61.8 & \colorbox{red!25}{16.7} & 0.347 & \colorbox{red!25}{73.5} & 73.3 & \colorbox{red!25}{51.4} \\ 
Prop. w/o SE\textsuperscript{\ding{61}} & 85.27 & 69.76 & 1.95 & 0.137 & 22.5 & 47.7 & 61.8 & 16.2 & 0.350 & 74.0 & 73.1 & \colorbox{red!25}{51.3} \\
Prop. w/o AA\textsuperscript{\ding{61}} & \colorbox{red!25}{84.97} & 69.41 & \colorbox{green!25}{1.70} & \colorbox{red!25}{0.133} & \colorbox{red!25}{20.7} & \colorbox{red!25}{46.1} & \colorbox{red!25}{60.8} & \colorbox{red!25}{17.2} & \colorbox{red!25}{0.332} & \colorbox{green!25}{74.4} & \colorbox{red!25}{72.0} & \colorbox{red!25}{51.3} \\
\hline
\multicolumn{13}{|l|}{\textbf{\textit{\large Training Strategy}}}\\
Prop. w/o MD\textsuperscript{\ding{61}} & \colorbox{red!25}{84.66} & \colorbox{red!25}{68.64} & \colorbox{red!25}{2.23} &\colorbox{red!25}{0.134} & 22.4 & \colorbox{red!25}{47.3} & 62.2 & \colorbox{red!25}{16.7} & 0.346 & \colorbox{red!25}{72.3} & \colorbox{red!25}{71.7} & \colorbox{red!25}{50.1} \\
Prop. w/o MT\textsuperscript{\ding{104}} & X & X & X &0.140 & 22.4 & 48.1 & 62.8 & 15.9 & 0.350 & \colorbox{green!25}{74.1} & \colorbox{green!25}{73.9} & \colorbox{green!25}{52.8} \\
Prop. w/o MHL\textsuperscript{\ding{61}} & \colorbox{green!25}{85.60} & 69.52 & 1.91 & \colorbox{red!25}{0.133} & \colorbox{red!25}{21.5} & \colorbox{red!25}{46.7} & \colorbox{red!25}{61.0} & \colorbox{red!25}{16.8} & \colorbox{red!25}{0.338} & \colorbox{red!25}{73.2} & 73.0 & \colorbox{red!25}{50.9} \\
\hline
Prop.\textsuperscript{\ding{61}} & 85.29 & 70.07 & 1.95 & 0.139 & 22.15 & 48.1 & 62.3 & 16.2 & 0.348 & 73.8 & 73.3 & 51.8 \\
SD(Prop.) & 0.2079 & 0.8026 & 0.1330 & 0.0046 & 0.64 & 0.56 & 0.88 & 0.2754 & 0.0050 & 0.2582 & 0.3594 & 0.3304 \\
\hline

\end{tabularx}
\caption{Results of the SIMMC-Fashion dataset, the \textbf{Prop.} is our proposed full model. All the details of the model descriptions can be referred to Table \ref{tab:result_furniture}}
\label{tab:result_fashion}
\end{table*}

\section{Experiments}
For the experiment, we use the SIMMC dataset and its evaluation metrics provided by the Facebook AI Research team in the DSTC9 track4. The datasets include three sub-tasks: \textbf{Sub-task 1 -- Multimodal Action Prediction}, \textbf{Sub-task 2 -- Multimodal Dialog Response Generation \& Retrieval}, \textbf{Sub-task 3 -- Multimodal Dialog State Tracking (MM-DST)}. 
In this section, we spell out the evaluation metrics, experiment details, and compare the performance between the baseline models \cite{moon2020situated} and our proposed model for all three sub-tasks.

\subsection{Experiment Details}
We adopt the \texttt{GPT2-small} model as the basis of all experiment, using the open-source implementation and pre-trained weights from \citet{Wolf2019HuggingFacesTS}.

In the training phase, we used the AdamW optimizer \cite{loshchilov2017decoupled} with $\beta_1$ = 0.9, $\beta_2$ = 0.999, and a learning rate of 5e-5 with linear decay. We fine-tuned the model with batch size 8, we also tried batch size 4,16 and 32 but did not notice significant difference on performance. 
For the loss function, we used CrossEntorpyLoss for all sub-tasks except for the Fashion API attributes prediction, where we used BCEWithLogitLoss since it is a multi-label classification task. 

For multi-domain learning, we merge the SIMMC-Fashion and SIMMC-Furniture datasets together as our training set. For each iteration, we randomly choose one domain and samples a batch from that domain, if we have exhausted one epoch for a domain, we keep sampling the remaining domain until it also iterated over one epoch.

In multi-task learning settings with API prediction task, we first train a GPT-2 LM head model on belief state and response with 6 epochs, and extend the model with multiple classifiers described in section \ref{sec:Model-Structure} and continue training 20 epochs.
For the first 2 epochs (after the 6 epochs) and the last 1 epoch, we trained the API action tasks, API attribute task, and LM task with a one-third chance for each. For the other training epochs, we only trained the API attribute task and LM task, with one-third and two-third chance, respectively.
 This scheduling process is designed based on the convergence speed of each task. API action task converges the first, API attribute prediction converges later and LM task converges the last. In multi-task learning, the total training epoch for API action is 3 and we found the best overall score for API attribute prediction and LM task (belief state and response) is between 12~14 epochs.  
 
%


During inference, instead of nucleus sampling used in baseline, we used Greedy decoding, which is more efficient and shown to achieve better performance in certain tasks\cite{ham-etal-2020-end}, and we also had the same observation in our early experiment.

\subsection{Evaluation Metrics}\label{ssec:evaluation}
The performance is evaluated on the dev-test split of the SIMMC dataset. We evaluate all sub-tasks on every turns of each dialogue, following the \textbf{Challenge Phase 1} settings of DSTC9 Track 4.
\paragraph{Sub-task 1 -- Multimodal Action Prediction}
In this sub-task, we evaluate the performance on API action, API action perplexity, and API attributes.
The API action is a multi-class classification task, with seven and five different actions for Furniture and Fashion domains, respectively. We evaluate the performance by accuracy and perplexity.
The accuracy of the API action is calculated by the ratio of the correct predictions in all predictions. The perplexity is defined as the exponential of the mean log-likelihood of the ground truth action in each turn.
For the API attributes prediction, in the Furniture domain, since it is a single-label classification problem, the performance is evaluated using accuracy. 
In the Fashion domain, the attributes prediction is a multi-label classification task. Thus it is evaluated using F1 score metrics.
\paragraph{Sub-task 2 -- Multimodal Dialog Response Generation \& Retrieval}
In sub-task 2, it is divided into two parts -- Generation score and Retrieval score. For the generation , \textbf{BLEU-4}~\cite{papineni2002bleu} is used as the evaluation metrics. For retrieval, a model needs to rank the given 100 candidate responses containing one GT (ground truth) response, then the retrieval score is calculated using several metrics: \textbf{Recall@1, Recall@5, Recall@10, Mean Reciprocal Rank (MRR)}, and \textbf{Mean Rank}. 

\begin{equation}
    \label{eq4}
    \begin{split}
    & \text{Greater}(a,b) = 
    \begin{cases} 
    1, & \text{if} \quad a>b \\
    0, & \text{otherwise}
    \end{cases} 
    \\
    & \text{Recall}@K = \frac{1}{N}\sum_{i=1}^N \text{Greater}(K, \text{Rank}^{GT}_i) \\
    & \text{Mean Rank} = \frac{1}{N}\sum_{i=1}^N (\text{Rank}^{GT}_i+1) \\
    & \text{MRR} = \frac{1}{N}\sum_{i=1}^N \frac{1}{(\text{Rank}^{GT}_i+1)} \\
    \end{split}
\end{equation}
The $\text{Rank}^{GT}$ in (\ref{eq4}) is the ranking of the ground truth response predicted by the model.
For simplicity, we score the 100 candidates by its \textbf{BLEU-4} scores measured against the response generated by our proposed model, and rank them accordingly.
\paragraph{Sub-task 3 -- Multimodal Dialog State Tracking (MM-DST)} In sub-task 3, we evaluate the Intents and Slots in belief states using F1 metric.
We also evaluate the joint accuracy of the belief states, which indicates the percentage where both the intents and slots are correct.

\subsection{Compared Baseline}
We compare our proposed model with the Action Prediction Model proposed by \citet{moon2020situated} and the GPT-2 based baseline model proposed on the DSTC9 Track4 Challenge Page. The Action Prediction Model is able to do sub-task 1 (API) and sub-task 2 (System Response), where we use the score release by the Facebook AI Research. The GPT-2 based baseline model is able to do sub-task 2 (System Response) and sub-task 3(Belief States), the details of these sub-tasks have been described in the prior of this section and section \ref{sec:Dataset}. For the GPT-2 Baseline Model, we retrain the model with more epochs (matching ours) and acquire better results compare to the scores Facebook AI Research has released, and use this score for comparison.


\subsection{Results}
We evaluated our proposed model on the dev-test split in SIMMC dataset. For Furniture and Fashion domain, the results are shown in Table \ref{tab:result_furniture} and Table \ref{tab:result_fashion}, respectively. In the tables, it can be seen that our proposed model outperforms the two baseline models with a great margin on all sub-tasks in both domains, except for the API Action of Furniture domain, which we believe has reached the upper bound of the performance, due to the noise in the datasets (For a single turn, more than one actions are reasonable answers). For example, the actions "None" and "SearchFurniture" are frequently both reasonable answers for a single turn.


\paragraph{Ablations}
To test the importance of the proposed features (the modification of inputs and training strategies, in Section \ref{sec:training_strategy} and Subsection \ref{ssec:input}), we conduct ablation studies on our proposed model.
In Table \ref{tab:result_furniture} and \ref{tab:result_fashion}, the proposed model scores and standard deviations are calculated based on repeatedly training four full models using different random seeds. Our proposed model demonstrates the best overall performance on all sub-tasks across domains.

First, we see that using more natural tokens with \textbf{Split Belief Intent} results in a slight increase in performance for the belief state on both domains.
When removing the \textbf{Segment Embedding}, the performance doesn't change much in both domains.
We can also observe that the response scores is significantly lower without \textbf{Adding Action to Inputs} in both the Furniture and Fashion domain, showing our model is able to ground the response on action information. In the Fashion domain, the removal of \textbf{Adding Action to Inputs} also leads to belief states performance drop.
Next, we observe that \textbf{Multi-Domain Training} improves the overall performance in the lower-resource domain (Fashion). For the Furniture domain, the API and Belief states also benefit from the multi-domain training.
As the \textbf{Multi-Task Training}, it allows us to extend the same model to other tasks with comparable or better performance. 
\textbf{Mask History Loss} also consistently improves the performance across generation tasks and overall benefits classification tasks on the Fashion domain.
In conclusion, the full model with all proposed features does have the best average performance on all sub-tasks and is able to predict all sub-tasks and both domains using one single model, showing the great efficacy of all our proposed features. Note that \citet{moon2020situated} have shown that using \textbf{Visual Information} results in an overall increase in performance, thus we didn't conduct ablation on visual (multi-modal) information.

\section{Discussion}\label{sec:discuss}


\subsection{Scaling}
We found using a larger turn number $T$ (using more turns of history as input) will significantly increase performance. Similarly, we found using larger model (e.g. \texttt{gpt2-medium}) can also boost performance significantly.
However, due to the limitation of our computing resources, we use turn number $T=2$ and the small model \texttt{gpt2-small} for all of our experiments.


\subsection{End-to-end multi-task training v.s. T5-style multi-task learning}
In our end-to-end multi-task training, we train the model to predict multiple targets sequentially.
In T5-style multi-task learning \cite{raffel2020exploring}, a model learns to predict either belief state or response, given the same context concatenated with a task-specific token as prefix/suffix.
One might expect the sequential training can help the latter task. We found they yield similar performance.
The reason why training to predict sequentially does not improve performance is likely because the earlier generated target contains only information extracted from the context, therefore, it doesn't benefit the latter tasks at inference time.
Despite that, the training process of our method is roughly two times faster than those of T5-style training. It is because each example in our method is equivalent to two examples in a T5-style training. 

\subsection{Using text-to-text generation for action prediction}\label{ssec:text_gen}
In our initial experiments, using text generation to predict API action and attributes couldn't surpass baseline performance. However, as suggested by anonymous reviewers, we revisited the idea with some of the proposed features (AA, MHL) and were able to achieve comparable performance on Action accuracy and Attribute accuracy. This result extends previous work \cite{raffel2020exploring} and shows text generation can be used for complex (2 stages and multi-label) classification tasks and achieve competitive performance. Notably, we achieved an exceptional $\approx$ 79\% accuracy on Fashion Attributes. We speculate that it is because fashion API has a more natural-looking text and thus benefits from the generation ability of the language model. Recent work \cite{schick2020its} also showed that using different patterns to cast the same classification task into text generation could result in a significant difference in performance. We leave further investigation to future work.

\section{Conclusion}
In this paper, we leveraged the iterative training technique from multi-task training to train multiple tasks and multi-domain data on one single model. By applying other proposed features, our model can utilize multi-modal information and outperforms the baseline models with a great margin. Our ablation studies showed that most of our proposed features contribute to the final performance. We hope that this simple yet effective framework and our ablation studies can serve as a reference for future researches.

\bibliography{citation.bib}
\end{document}